\documentclass[10pt,twocolumn,letterpaper,table]{article}

\usepackage{cvpr}
\usepackage{times}
\usepackage{epsfig}
\usepackage{graphicx}
\usepackage{amsmath}

\usepackage{amssymb}
\usepackage{gensymb}
\usepackage{comment}
\usepackage{multirow}
\usepackage{subcaption}
\usepackage{booktabs}
\usepackage{microtype}
\usepackage{caption}
\usepackage{floatrow}
\newfloatcommand{capbtabbox}{table}[][\FBwidth]

\usepackage{colortbl}
\usepackage{booktabs}
\usepackage{xfrac}
\usepackage{tablefootnote}
\usepackage{tabularx}
\definecolor{rowblue}{RGB}{220,230,240}

\newcommand{\eucsqnorm}[1]{\left\|{#1}\right\|_2^2}
\newcommand{\absnorm}[1]{\left\|{#1}\right\|_1}

\newcommand{\I}{\mathcal{I}}

\newcommand{\Trans}{\mathcal{T}}

\newcommand{\p}{\mathbf{p}}

\newcommand{\trans}{\mathbf{t}}
\newcommand{\0}{\mathbf{0}}

\newcommand{\x}{\mathbf{x}}

\newcommand{\G}{\mathcal{G}}

\newcommand{\Real}{\mathbb{R}}

\newcommand{\loss}{\mathcal{L}}

\newcommand{\mesh}{\mathcal{M}}
\newcommand{\proj}{\pi}
\newcommand{\iproj}{\proj^{-1}}
\newcommand{\projf}[1]{\proj\left({#1}\right)}
\newcommand{\iprojf}[1]{\iproj\left({#1}\right)}
\newcommand{\cam}{\boldsymbol{\Omega}}
\newcommand{\vtx}{\mathbf{v}}
\newcommand{\baryc}{\boldsymbol{\alpha}}
\newcommand{\V}{\mathbf{V}}

\newcommand{\SFM}{S\emph{f}M\xspace}

\newcommand{\VT}{\text{V}}
\newcommand{\ph}{\text{photo}}
\newcommand{\btheta}{\boldsymbol{\theta}}

\newcommand{\bomega}{\boldsymbol{\omega}}

\newcommand{\z}{\mathbf{z}}
\newcommand{\set}{\mathcal{S}}
\newcommand{\Pset}{\mathcal{P}}

\newcommand{\Rot}{\mathcal{R}}

\usepackage[pagebackref=true,breaklinks=true,letterpaper=true,colorlinks,bookmarks=false]{hyperref}

\cvprfinalcopy %

\def\cvprPaperID{787} %

\ifcvprfinal\fi
\begin{document}

\title{Photometric Mesh Optimization for Video-Aligned 3D Object Reconstruction}

\author{Chen-Hsuan Lin\textsuperscript{1,2\ref{fn}} \quad
Oliver Wang\textsuperscript{2} \quad
Bryan C. Russell\textsuperscript{2} \quad
Eli Shechtman\textsuperscript{2} \\
Vladimir G. Kim\textsuperscript{2} \quad
Matthew Fisher\textsuperscript{2} \quad
Simon Lucey\textsuperscript{1} \vspace{4pt} \\ 
\textsuperscript{1}The Robotics Institute, Carnegie Mellon University \qquad
\textsuperscript{2}Adobe Research \vspace{-2pt} \\
{\tt\small chlin@cmu.edu \quad \{owang,brussell,elishe,vokim,matfishe\}@adobe.com \quad slucey@cs.cmu.edu} \vspace{2pt} \\
{\small \url{https://chenhsuanlin.bitbucket.io/photometric-mesh-optim/}}
}

\maketitle
\renewcommand*{\thefootnote}{\fnsymbol{footnote}}
\setcounter{footnote}{1}
\footnotetext{Work done during CHL's internship at Adobe Research.\label{fn}}
\renewcommand*{\thefootnote}{\arabic{footnote}}
\setcounter{footnote}{0}

\begin{abstract}

In this paper, we address the problem of 3D object mesh reconstruction from RGB videos.
Our approach combines the best of multi-view geometric and data-driven methods for 3D reconstruction by optimizing object meshes for multi-view photometric consistency while constraining mesh deformations with a shape prior.
We pose this as a piecewise image alignment problem for each mesh face projection.
Our approach allows us to update shape parameters from the photometric error without any depth or mask information.
Moreover, we show how to avoid a degeneracy of zero photometric gradients via rasterizing from a virtual viewpoint.
We demonstrate 3D object mesh reconstruction results from both synthetic and real-world videos with our photometric mesh optimization, which is unachievable with either na\"ive mesh generation networks or traditional pipelines of surface reconstruction without heavy manual post-processing.

\end{abstract}

\section{Introduction}

The choice of 3D representation plays a crucial role in 3D reconstruction problems from 2D images.
Classical multi-view geometric methods, most notably structure from motion (\SFM) and SLAM, recover point clouds as the underlying 3D structure of RGB sequences, often with very high accuracy~\cite{engel2014lsd,newcombe2011dtam}.
Point clouds, however, lack inherent 3D spatial structure that is essential for efficient reasoning.
In many scenarios, \emph{mesh representations} are more desirable -- they are significantly more compact since they have inherent geometric structures defined by point connectivity, while they also represent continuous surfaces necessary for many applications such as robotics (\eg, accurate localization for autonomous driving), computer graphics (\eg, physical simulation, texture synthesis), and virtual/augmented reality.

Another drawback of classical multi-view geometric methods is reliance on hand-designed features and can be fragile when their assumptions are violated. 
This happens especially in textureless regions or when there are changes in illumination.
Data-driven approaches~\cite{choy20163d,groueix2018atlasnet}, on the other hand, learn priors to tackle ill-posed 3D reconstruction problems and have recently been widely applied to 3D prediction tasks from single images.
However, they can only reliably reconstruct from the space of training examples it learns from, resulting in limited ability to generalize to unseen data.

\begin{figure}[t!]
	\centering
	\includegraphics[width=1.0\linewidth]{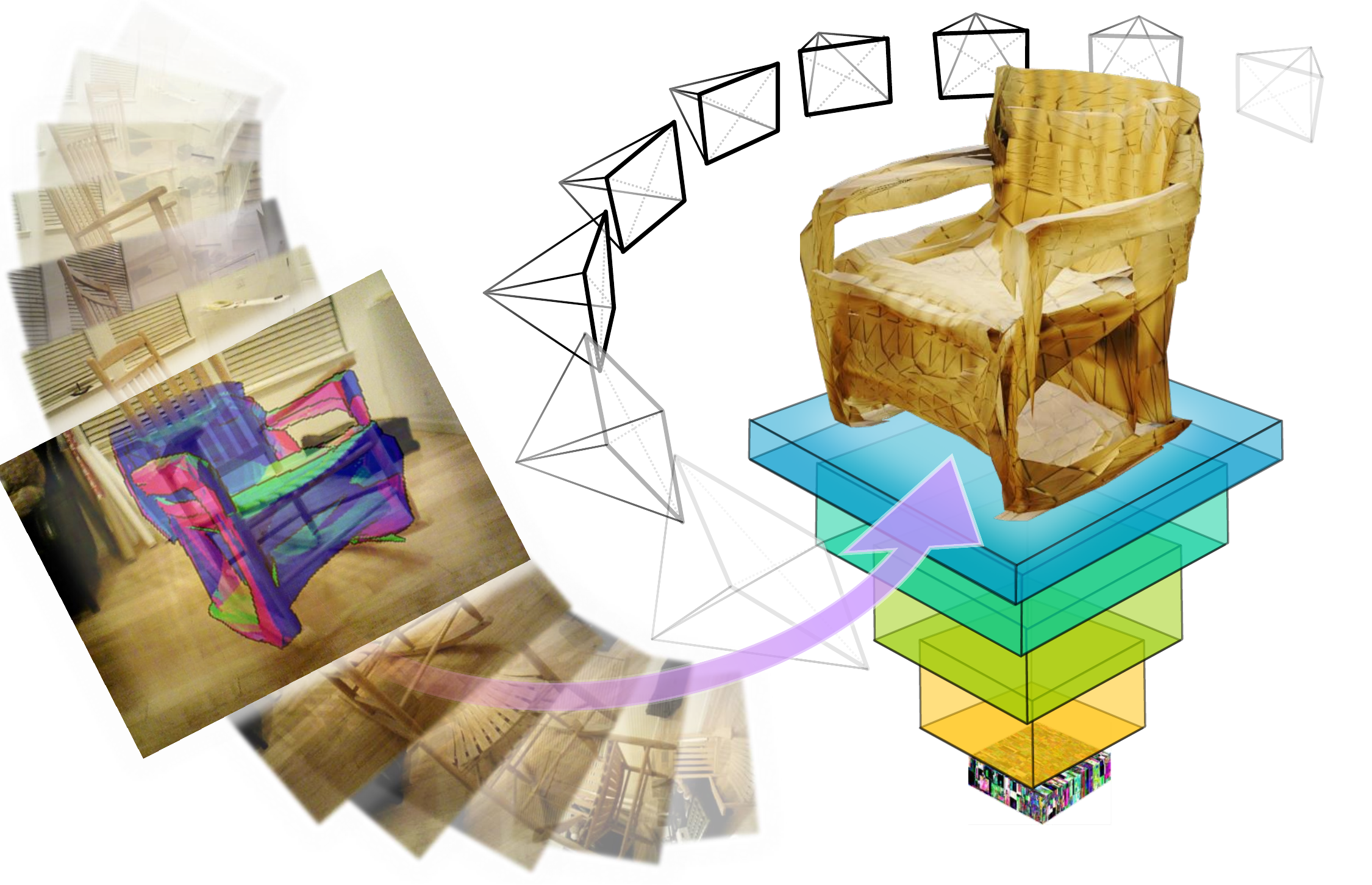}
	\caption{Our video-aligned object mesh reconstruction enforcing multi-view consistency while constraining shape deformations with shape priors, generating an output mesh with improved geometry with respect to the input views.}
	\label{fig:teaser}
\end{figure}

In this work, we address the problem of 3D mesh reconstruction from image sequences by bringing together the best attributes of multi-view geometric methods and data-driven approaches (Fig.~\ref{fig:teaser}).
Focusing on object instances, we use \emph{shape priors} (specifically, neural networks) to reconstruct geometry with incomplete observations as well as \emph{multi-view geometric constraints} to refine mesh predictions on the input sequences.
Our approach allows dense reconstruction with object semantics from learned priors, which is not possible from the traditional pipelines of surface meshing~\cite{kazhdan2013screened} from multi-view stereo (MVS).
Moreover, our approach generalizes to unseen objects by utilizing multi-view geometry to enforce observation consistency across viewpoints.

Given only RGB information, we achieve mesh reconstruction from image sequences by photometric optimization, which we pose as a piecewise image alignment problem of individual mesh faces.
To avoid degeneracy, we introduce a novel virtual viewpoint rasterization to compute photometric gradients with respect to mesh vertices for 3D alignment, allowing the mesh to deform to the observed shape.
A main advantage of our photometric mesh optimization is its non-reliance on any a-priori known depth or mask information~\cite{cmrKanazawa18,drcTulsiani17,yan2016perspective} -- a necessary condition to be able to reconstruct objects from real-world images.
With this, we take a step toward practical usage of prior-based 3D mesh reconstruction aligned with RGB sequences. 

In summary, we present the following contributions:
\begin{itemize}
	\setlength\itemsep{0px}
	\item We incorporate multi-view photometric consistency with data-driven shape priors for optimizing 3D meshes using 2D photometric cues.
	\item We propose a novel photometric optimization formulation for meshes and introduce a virtual viewpoint rasterization step to avoid gradient degeneracy.
\end{itemize}
Finally, we show 3D object mesh reconstruction results from both synthetic and real-world sequences, unachievable with either na\"ive mesh generators or traditional MVS pipelines without heavy manual post-processing.

\section{Related Work}

Our work on object mesh reconstruction touches several areas, including multi-view object reconstruction, mesh optimization, deep shape priors, and image alignment.

\vspace{6px}
\noindent\textbf{Multi-view object reconstruction.} Multi-view calibration and reconstruction is a well-studied problem.
Most approaches begin by estimating camera coordinates using 2D keypoint matching, a process known as SLAM~\cite{engel2014lsd,mur2015orb} or SfM~\cite{fuentes2015visual,schoenberger2016sfm}, followed by dense reconstruction methods such as MVS~\cite{furukawa2010accurate} and meshing~\cite{kazhdan2013screened}.
More recent works using deep learning have explored 3D reconstruction from multiple-view consistency between various forms of 2D observations~\cite{lin2018learning,mvcTulsiani18,drcTulsiani17,yan2016perspective,zhu2017rethinking}.
These methods all utilize forms of 2D supervision that are easier to acquire than 3D CAD models, which are relatively limited in quantity.
Our approach uses both geometric and image-based constraints, which allows it to overcome common multi-view limitations such as missing observations and textureless regions. 

\vspace{6px}
\noindent\textbf{Mesh optimization.}
Mesh optimization dates back to classical works of Active Shape Models~\cite{cootes1992active} and Active Appearance Models~\cite{cootes2001active,matthews2004active}, which uses 2D meshes to fit facial landmarks.
In this work, we optimize for 3D meshes using 2D photometric cues, a significantly more challenging problem due to the inherent ambiguities in the task.
Similar approaches for mesh refinement have also been explored~\cite{delaunoy2014photometric,delaunoy2011gradient}; however, a sufficiently good initialization is required with very small vertex perturbations allowed.
As we show in our experiments, we are able to handle larger amount of noise perturbation by optimizing over a latent shape code instead of mesh vertices, making it more suitable for practical uses.

Several recent methods have addressed learning 3D reconstruction with mesh representations.
AtlasNet~\cite{groueix2018atlasnet} and Pixel2Mesh~\cite{wang2018pixel2mesh} are examples of learning mesh object reconstructions from 3D CAD models.
Meanwhile, Neural Mesh Renderer~\cite{kato2018neural} suggested a method of mesh reconstruction via approximate gradients for 2D mask optimization, and Kanazawa~\etal~\cite{cmrKanazawa18} further advocated learning mesh reconstruction from 2D supervision of textures, masks, and 2D keypoints.
Our approach, in contrast, does \emph{not} assume any availability of masks or keypoints and operates purely via photometric cues across viewpoints.

\vspace{6px}
\noindent\textbf{Shape priors.}
The use of neural networks as object priors for reconstruction has recently been explored with point clouds~\cite{zhu2018object}.
However, it requires object masks as additional constraints during optimization.
We eliminate the need for mask supervision by regularizing the latent code.
Shape priors have also been explored for finding shape correspondences~\cite{groueix20183d}, where the network learns the deformation field from a template shape to match 3D observations.
In our method, we directly optimize the latent shape code to match 2D cues from multiple viewpoints and do not require a known shape template for the object.
A plane and primitive prior has been used for the challenging task of multi-view scene reconstruction~\cite{huang2017dlite}.
Although the primitive prior does not need to be learned from an object dataset, the resulting reconstruction can differ significantly from the target geometry when it is not well represented by the chosen primitives.

\vspace{6px}
\noindent\textbf{Image alignment.}
The most generic form of image alignment refers to prediction of inherent geometric misalignment between a pair of images.
Image alignment using simple warping functions can be dated back to the seminal Lucas-Kanade algorithm~\cite{lucas1981iterative} and its recent variants~\cite{baker2004lucas,lin2016conditional}.
Recent work has also explored learning a warp function to align images from neural networks for applications such as novel view synthesis~\cite{zhou2017unsupervised,zhou2016view} and learning invariant representations~\cite{jaderberg2015spatial,lin2017inverse}.
In this work, we pose our problem of mesh optimization as multiple image alignment problems of mesh faces, and solve it by optimizing over a latent code from a deep network rather than the vertices themselves.

\section{Approach} \label{sec:approach}

\begin{figure}[t!]
	\centering
	\includegraphics[width=1.0\linewidth]{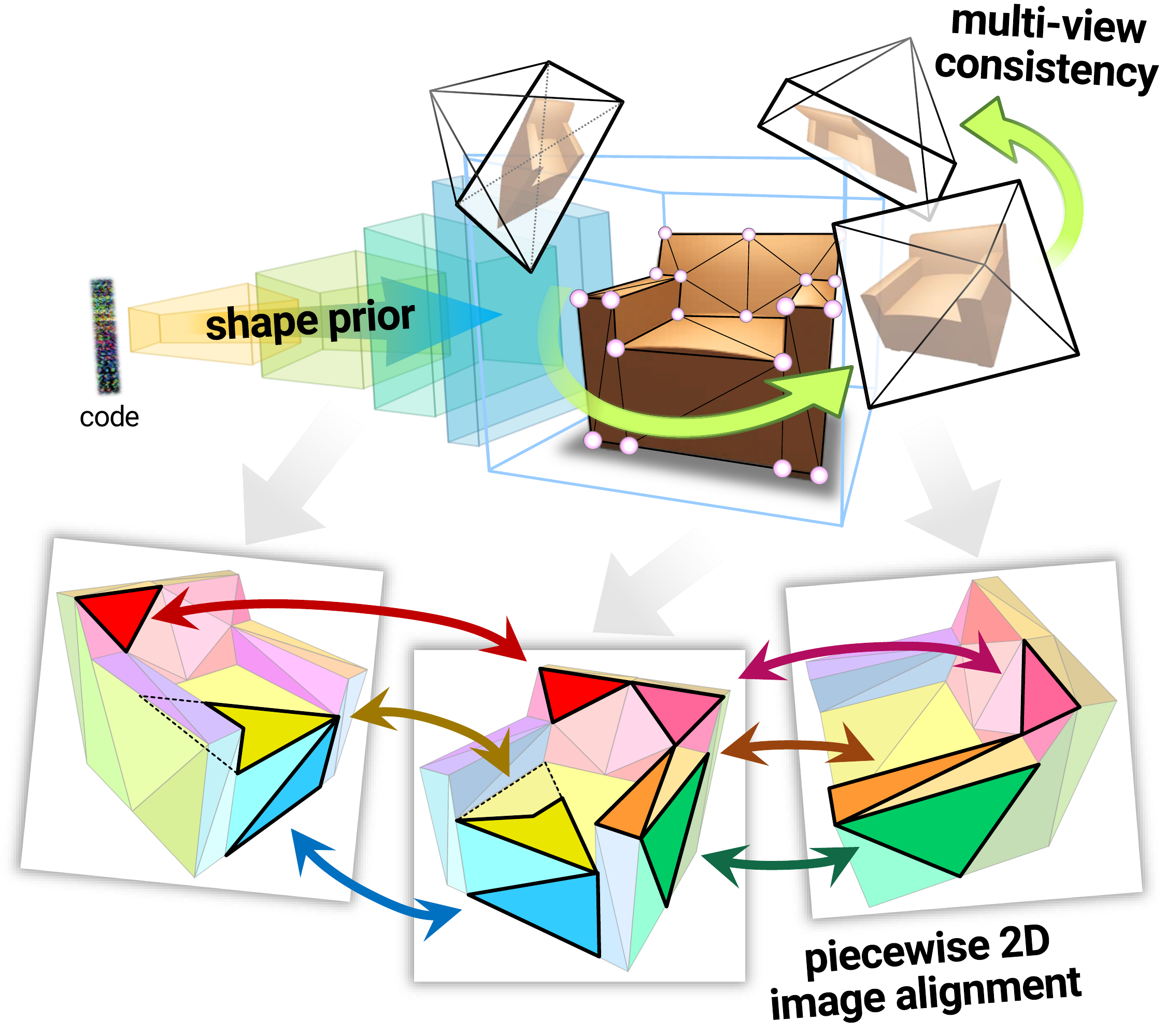}
	\caption{Overview.
	We perform 3D mesh reconstruction via \emph{piecewise image alignment} of triangles to achieve per-triangle visibility-aware photometric consistency across multiple views, with mesh vertices optimized over the latent code of a shape prior learned by deep neural networks.
	}
	\label{fig:overview}
\end{figure}

We seek to reconstruct a 3D object mesh from an RGB sequence $\{(\I_f,\cam_f)\}$, where each frame $\I_f$ is associated with a camera matrix $\cam_f$.
In this work, we assume that
the camera matrices $\{\cam_{\hspace{-1pt}f}\}$ can be readily obtained from off-the-shelf \SFM methods~\cite{schoenberger2016sfm}.
Fig.~\ref{fig:overview} provides an overview -- we optimize for object meshes that maximize multi-view photometric consistency over a shape prior, where we use a pretrained mesh generator.
We focus on triangular meshes here although our method is applicable to any mesh type.


\subsection{Mesh Optimization over Shape Prior} \label{sec:optim}

Direct optimization on a 3D mesh $\mesh$ with $N$ vertices involves solving for $3N$ degrees of freedom (DoFs) and typically becomes underconstrained when $N$ is large.
Therefore, reducing the allowed DoFs is crucial to ensure mesh deformations are well-behaved during optimization.
We wish to represent the mesh $\mesh = \G(\z)$ as a differentiable function $\G$ of a reduced vector representation $\z$.

We propose to use an off-the-shelf generative neural network as the main part of $\G$ and reparameterize the mesh with an associated latent code $\z \in \Real^{K}$, where $K\hspace{-1pt}\ll\hspace{-1pt}3N$.
The network serves as an object shape prior whose efficacy comes from pretraining on external shape datasets.
Shape priors over point clouds have been previously explored~\cite{zhu2018object}; here, we extend to mesh representations.
We use AtlasNet~\cite{groueix2018atlasnet} here although other mesh generators are also applicable.
The shape prior allows the predicted 3D mesh to deform within a learned shape space, avoiding many local minima that exist with direct vertex optimization.
To utilize RGB information from the given sequence for photometric optimization, we further add a 3D similarity transform to map the generated mesh to world cameras recovered by \SFM (see Sec.~\ref{sec:implementation}).

We define our optimization problem as follows: given the RGB image sequence and cameras $\{(\I_f,\cam_f)\}$, we optimize a regularized cost consisting of a photometric loss $\loss_\ph$ for all pairs of frames over the representation $\z$, formulated as
\begin{align} \label{eq:objective}
	\min_{\z} \sum_{ a \neq b } \loss_\ph(\I_a,\I_b,\cam_a,\cam_b;\G(\z)) + \loss_\text{reg}(\z) \;, 
\end{align}
where $\loss_\text{reg}$ is a regularization term on $\z$.
This objective allows the generated mesh to deform with respect to an effective shape prior.
We describe each term in detail next.


\subsection{Piecewise Image Alignment} \label{sec:alignment}

Optimizing the mesh $\mesh$ with the photometric loss $\loss_\ph$ is based on the assumption that a dense 2D projection of the individual triangular faces of a 3D mesh should be globally consistent across multiple viewpoints.
Therefore, we cast the problem of 3D mesh alignment to the input views as a collection of \emph{piecewise 2D image alignment} subproblems of each projected triangular face (Fig.~\ref{fig:overview}).

To perform piecewise 2D image alignment between $\I_a$ and $\I_b$, we need to establish pixel correspondences.
We first denote $\V_j(\z)\ \in \Real^{3 \times 3}$ as the 3D vertices of triangle $j$ in mesh $\mesh = \G(\z)$, defined as column vectors.
From triangle $j$, we can sample a collection of 3D points $\Pset_j = \{\p_i(\z)\}$ that lie within triangle $j$, related via $\p_i(\z) = \V_j(\z)\baryc_i$ through the barycentric coordinates $\baryc_i$.
For a camera $\cam$, let $\proj : \Real^3 \to \Real^2$ be the projection function mapping a world 3D point $\p_i(\z)$ to 2D image coordinates.
The pixel intensity error between the two views $\cam_a$ and $\cam_b$ can be compared at the 2D image coordinates corresponding to the projected sampled 3D points.
We formulate the photometric loss $\loss_\ph$ as the sum of $\ell_1$ distances between pixel intensities at these 2D image coordinates over all triangular faces,
\begin{align} \label{eq:photom}
	& \loss_\ph (\I_a,\I_b,\cam_a,\cam_b;\G(\z)) \\
	= & \sum_{j} \sum_{i: \p_i \in \Pset_j} \absnorm{ \I_a \left( \projf{ \p_i(\z); \cam_a } \right) - \I_b \left( \projf{ \p_i(\z); \cam_b } \right) } \; . \nonumber
\end{align}
As such, we can optimize the photometric loss $\loss_\ph$ with pixel correspondences established as a function of $\z$.

\vspace{6px}
\noindent\textbf{Visibility.}
As a 3D point $\p_i$ may not be visible in a given view due to possible object self-occlusion, we handle visibility by constraining $\Pset_j$ to be the set of samples in triangle $j$ whose projection is visible in both views.
We achieve this by returning a mesh index map using mesh rasterization, a standard operation in computer graphics, for each optimization step.
The photometric gradients of each sampled point $\frac{\partial \I}{\partial \V_j} = \frac{\partial \I}{\partial \x_i} \frac{\partial \x_i}{\partial \p_i} \frac{\partial \p_i}{\partial \V_j}$ in turn backpropagate to the vertices $\V_j$.
We obtain $\frac{\partial \I}{\partial \x_i}$ through differentiable image sampling~\cite{jaderberg2015spatial}, $\frac{\partial \x_i}{\partial \p_i}$ by taking the derivative of the projection $\proj$, and $\frac{\partial \p_i}{\partial \V_j}$ by associating with the barycentric coordinates $\baryc_i$.
We note that the entire process is differentiable and does not resort to approximate gradients~\cite{kato2018neural}.


\begin{figure}[t!]
	\centering
	\includegraphics[width=1.0\linewidth]{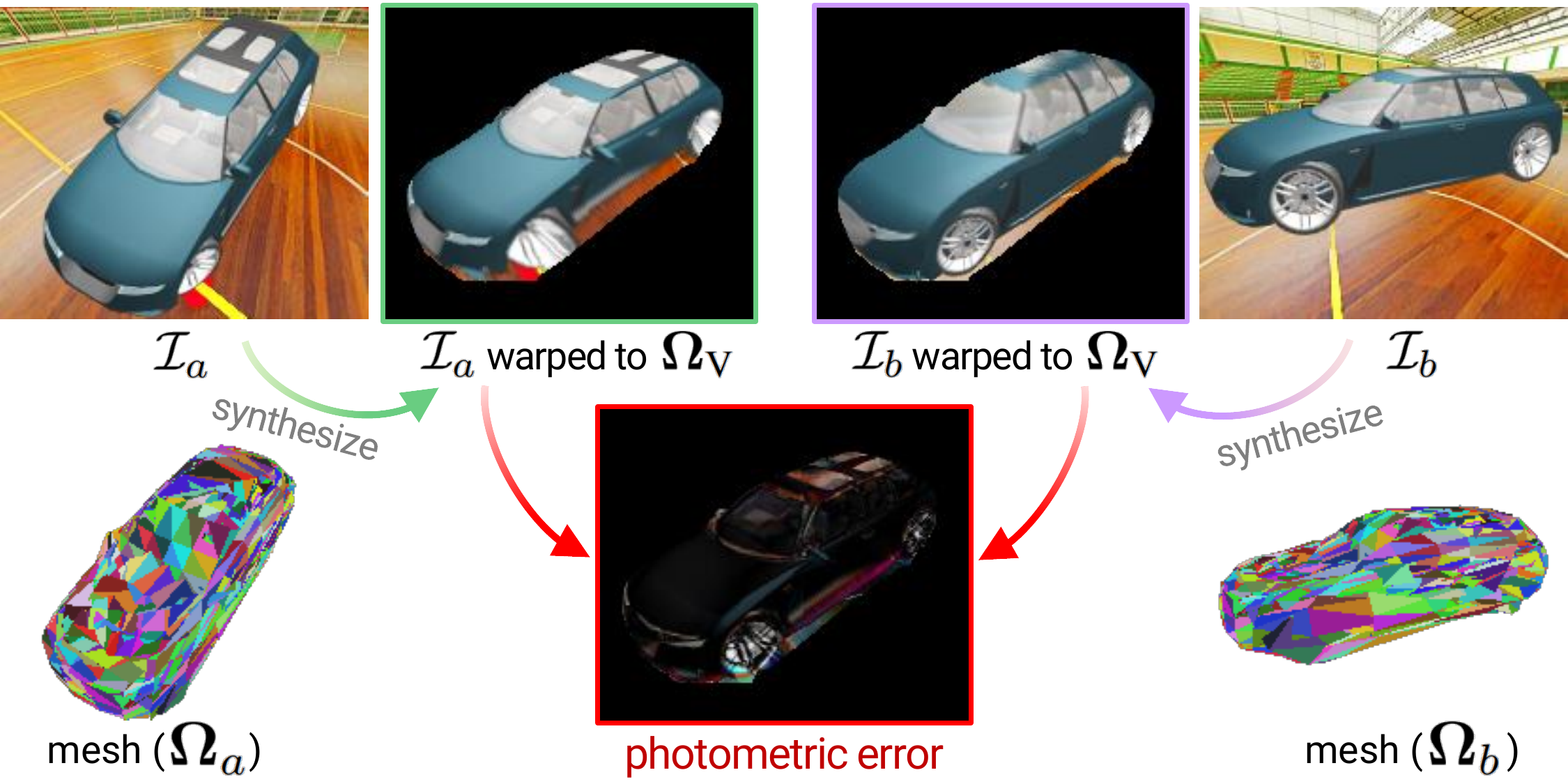}
	\caption{Visualization of the photometric loss $\loss_\ph$ between the synthesized appearances at virtual viewpoints $\cam_\VT$ starting from input images $\I_a$ and $\I_b$.
	The photometric loss $\loss_\ph$ encourages consistent appearance syntheses from both input viewpoints $\cam_a$ and $\cam_b$.
	}
	\label{fig:virtual}
\end{figure}

\subsection{Virtual Viewpoint Rasterization} \label{sec:virtual}

We can efficiently sample a large number of 3D points $\Pset_j$ in triangle $j$ by rendering the depth of $\mesh$ from a given view using mesh rasterization (Sec.~\ref{sec:alignment}).
If the depth were rasterized from either input view $\cam_a$ or $\cam_b$, however, we would obtain zero photometric gradients. %
This degeneracy arises due to the fact that ray-casting from one view and projecting back to the same view results in $\frac{\partial \I}{\partial \V_j} = \0$.

To elaborate, we first note that depth rasterization of triangle $j$ is equivalent to back-projecting regular grid coordinates $\bar{\x}_i$ to triangle $j$.
We can express each depth point from camera $\cam \in \{\cam_a,\cam_b\}$ as $\p_i(\z) = \iprojf{ \bar{\x}_i ; \V_j(\z),\cam }$, where $\iproj : \Real^2 \to \Real^3$ is the inverse projection function realized by solving for ray-triangle intersection with $\V_j(\z)$.
Combining with the projection equation, we have
\begin{align} \label{eq:degen}
	\x_i(\z,\cam) = \projf{ \iprojf{ \bar{\x}_i; \V_j(\z), \cam } ; \cam } = \bar{\x}_i \;\;\; \forall \bar{\x}_i \;,
\end{align}
becoming the identity mapping and losing the dependency of $\x_i$ on $\V_j(\z)$, which in turn leads to $\frac{\partial \x_i}{\partial \V_j} = \0$.
This insight is in line with the recent observation from Ham~\etal~\cite{ham2017proxy}.

To overcome this degeneracy, we rasterize the depth from a \emph{third} virtual viewpoint $\cam_\VT \notin \{\cam_a,\cam_b\}$.
This step allows correct gradients to be computed in both viewpoints $\cam_a$ and $\cam_b$, which is essential to maintain stability during optimization.
We can form the photometric loss by synthesizing the image appearance at $\cam_\VT$ using the pixel intensities from both $\cam_a$ and $\cam_b$ (Fig.~\ref{fig:virtual}).
We note that $\cam_\VT$ can be arbitrarily chosen.
In practice, we choose $\cam_\VT$ to be the bisection between $\cam_a$ and $\cam_b$ by applying Slerp~\cite{shoemake1985animating} on the rotation quaternions and averaging the two camera centers.


\subsection{Implementation Details} \label{sec:implementation}

\noindent\textbf{Coordinate systems.}
Mesh predictions from a generative network typically lie in a canonical coordinate system~\cite{groueix2018atlasnet,wang2018pixel2mesh} independent of the world cameras recovered by \SFM.
Therefore, we need to account for an additional 3D similarity transform $\Trans(\cdot)$ applied to the mesh vertices.
For each 3D vertex $\vtx_k'$ from the prediction, we define the similarity transform as 
\begin{align}
	\vtx_k = \Trans(\vtx_k';\btheta) = \exp(s) \cdot \Rot(\bomega) \vtx_k' + \trans \;\; \forall k \;,
\end{align}
where $\btheta = [s;\bomega;\trans] \in \Real^7$ are the parameters and $\Rot$ is a 3D rotation matrix parameterized with the $\mathfrak{so}(3)$ Lie algebra.
We optimize for $\z = [\z';\btheta]$ together, where $\z'$ is the latent code associated with the generative network.

Since automated registration of noisy 3D data with unknown scales is still an open problem, we assume a coarse alignment of the coordinate systems can be computed from minimal annotation of rough correspondences (see Sec.~\ref{sec:real} for details).
We optimize for the similarity transform to more accurately align the meshes to the RGB sequences.

\vspace{6px}
\noindent\textbf{Regularization.}
Despite neural networks being effective priors, the latent space is only spanned by the training data.
To avoid meshes from reaching a degenerate solution, we impose an extra penalty on the latent code $\z'$ to ensure it stays within a trust region of the initial code $\z_0$ (extracted from a pretrained image encoder), defined as $\loss_\text{code} = \eucsqnorm{\z'-\z_0}$. %
We also add a scale penalty $\loss_\text{scale} = -s$ that encourages the mesh to expand, since the mesh shrinking to infinitesimal is a trivial solution with zero photometric error.
The regularization $\loss_\text{reg}$ in cost~\eqref{eq:objective} is written as
\begin{align}
	\loss_\text{reg}(\z) = \lambda_\text{code}\cdot \loss_\text{code}(\z') + \lambda_\text{scale}\cdot \loss_\text{scale}(\btheta)
\end{align}
where $\lambda_\text{code}$ and $\lambda_\text{scale}$ are the penalty weights.

\section{Experiments}

We evaluate the performance of our method on a single (Sec.~\ref{sec:single}) and multiple (Sec.~\ref{sec:multi}) object categories with synthetic data as well as real-world videos (Sec.~\ref{sec:real}). 

\begin{figure}[t!]
	\centering
	\includegraphics[width=1.0\linewidth]{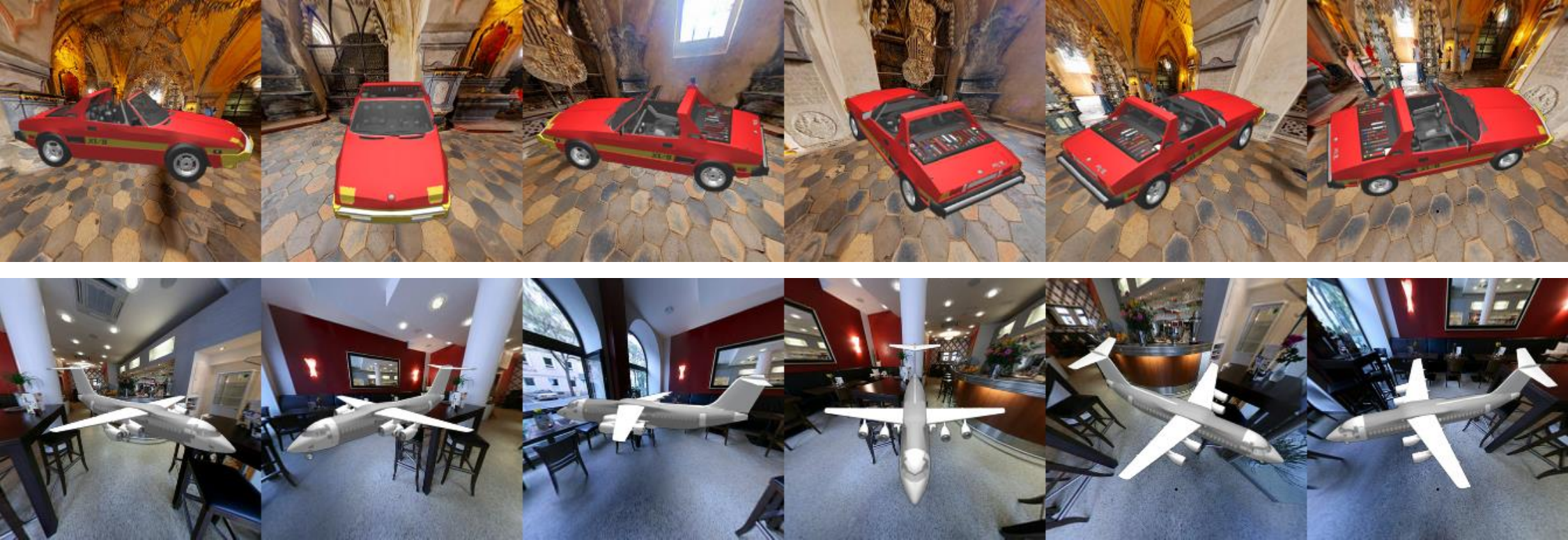}
	\caption{Sample sequences composited from ShapeNet renderings (top: car, bottom: airplane) and SUN360 scenes.}
	\label{fig:sampleseq}
\end{figure}

\vspace{6px}
\noindent\textbf{Data preparation.}
We create datasets of 3D CAD model renderings for training a mesh generation network and evaluating our optimization framework.
Our rendering pipeline aims to create realistic images with complex backgrounds so they could be applied to real-world video sequences. 
We use ShapeNet~\cite{chang2015shapenet} for the object dataset and normalize all objects to fit an origin-centered unit sphere.
We render RGB images of each object using perspective cameras at 24 equally spaced azimuth angles and 3 elevation angles. 

To simulate realistic backgrounds, we randomly warp and crop spherical images from the SUN360 database~\cite{xiao2012recognizing} to create background images of the same scene taken at different camera viewpoints.
By compositing the foreground and background images together at corresponding camera poses, we obtain RGB sequences of objects composited on realistic textured backgrounds (Fig.~\ref{fig:sampleseq}).
Note that we do not keep any mask information that was accessible in the rendering and compositing process as such information is typically not available in real-world examples.
All images are rendered/cropped at a resolution of 224$\times$224.

\vspace{6px}
\noindent\textbf{Shape prior.}
We use AtlasNet~\cite{groueix2018atlasnet} as the base network architecture for mesh generation, which we retrain on our new dataset.
We use the same 80\%-20\% training/test split from Groueix~\etal~\cite{groueix2018atlasnet} and additionally split the SUN360 spherical images with the same ratio.
During training, we augment background images at random azimuth angles.

\begin{figure*}[t!]
	\centering
	\includegraphics[width=1.0\linewidth]{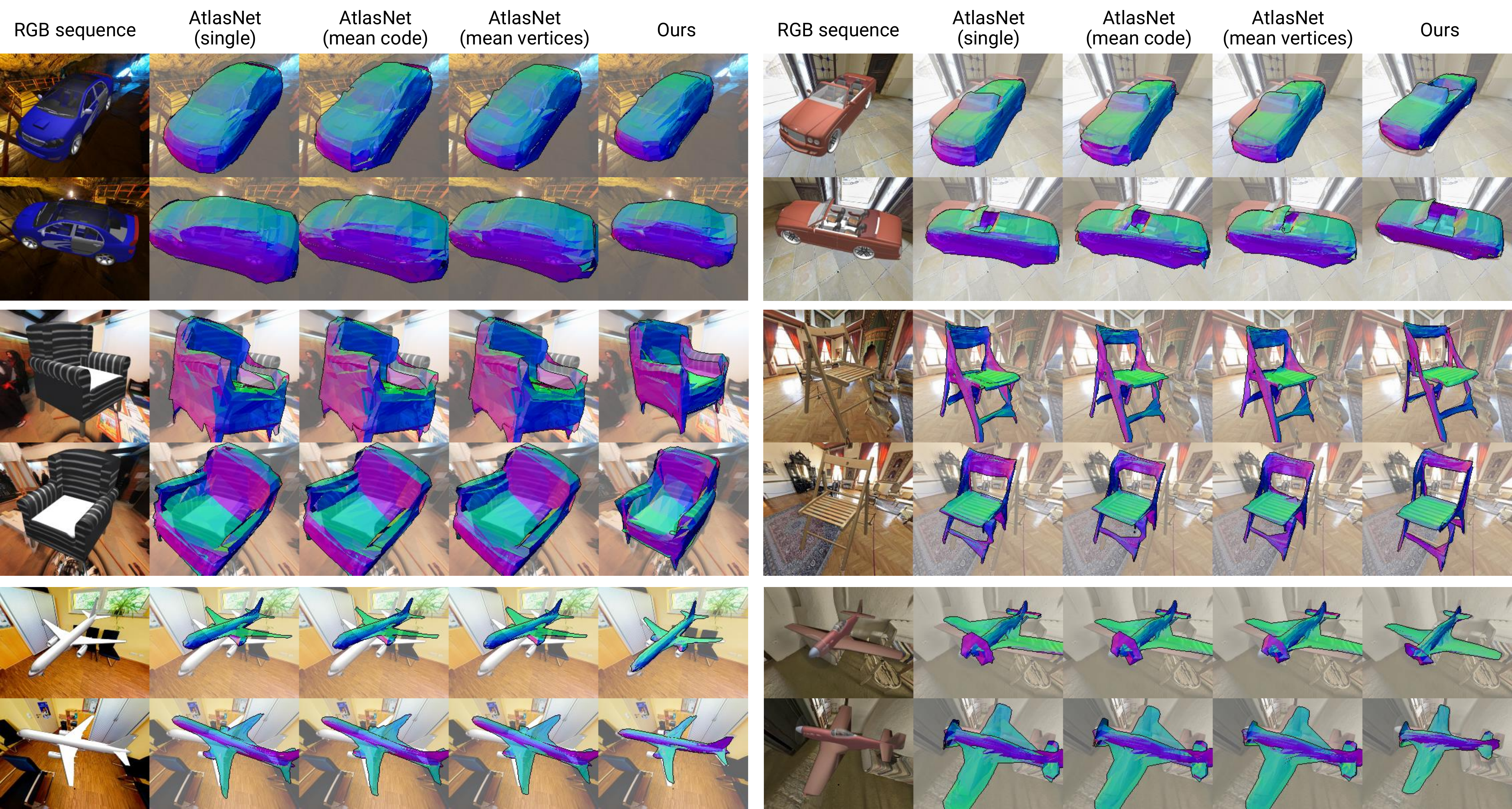}
	\caption{Qualitative results from category-specific models, where we visualize two sample frames from each test sequence.
	Our method better aligns initial meshes to the RGB sequences while optimizing for more subtle shape details (\eg, car spoilers and airplane wings) over baselines.
	The meshes are color-coded by surface normals with occlusion boundaries drawn.}
	\label{fig:results_single}
\end{figure*}

\vspace{6px}
\noindent\textbf{Initialization.}
We initialize the code $\z_0$ by encoding an RGB frame with the AtlasNet encoder.
For ShapeNet sequences, we choose frames with objects facing $45\degree$ sideways.
For real-world sequences, we manually select frames where objects are center-aligned to the images as much as possible to match our rendering settings.
We initialize the similarity transform parameters to $\btheta = \0$ (identity transform).

\vspace{6px}
\noindent\textbf{Evaluation criteria.}
We evaluate the result by measuring the 3D distances between the sampled 3D points from the predicted meshes and the ground-truth point clouds~\cite{groueix2018atlasnet}.
We follow Lin~\etal~\cite{lin2018learning} by reporting the 3D error between the predicted and ground-truth point clouds as $\eta (\set_1,\set_2) = \sum_{i: \vtx_i \in \set_1} \min_{\vtx_j \in \set_2} \| \vtx_i-\vtx_j \|_2$ for some source and target point sets $\set_1$ and $\set_2$, respectively.
This metric measures the prediction shape accuracy when $\set_1$ is the prediction and $\set_2$ is the ground truth, while it indicates the prediction shape coverage when vice versa.
We report quantitative results in both directions separately averaged across all instances.


\subsection{Single Object Category} \label{sec:single}

\begin{figure}[t!]
	\centering
	\includegraphics[width=1.0\linewidth]{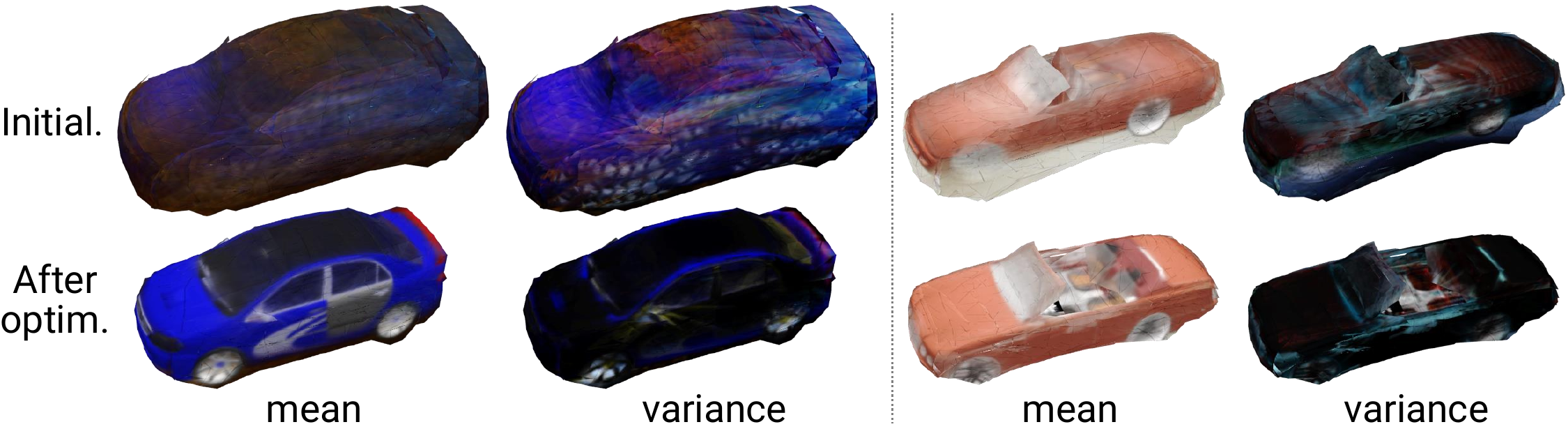}
	\caption{Mesh visualization with textures computed by averaging projections across all viewpoints.
	Our method successfully reduces variance and recovers dense textures that can be embedded on the surfaces.}
	\label{fig:single_mesh}
\end{figure}

\begin{figure}[t!]
	\centering
	\includegraphics[width=1.0\linewidth]{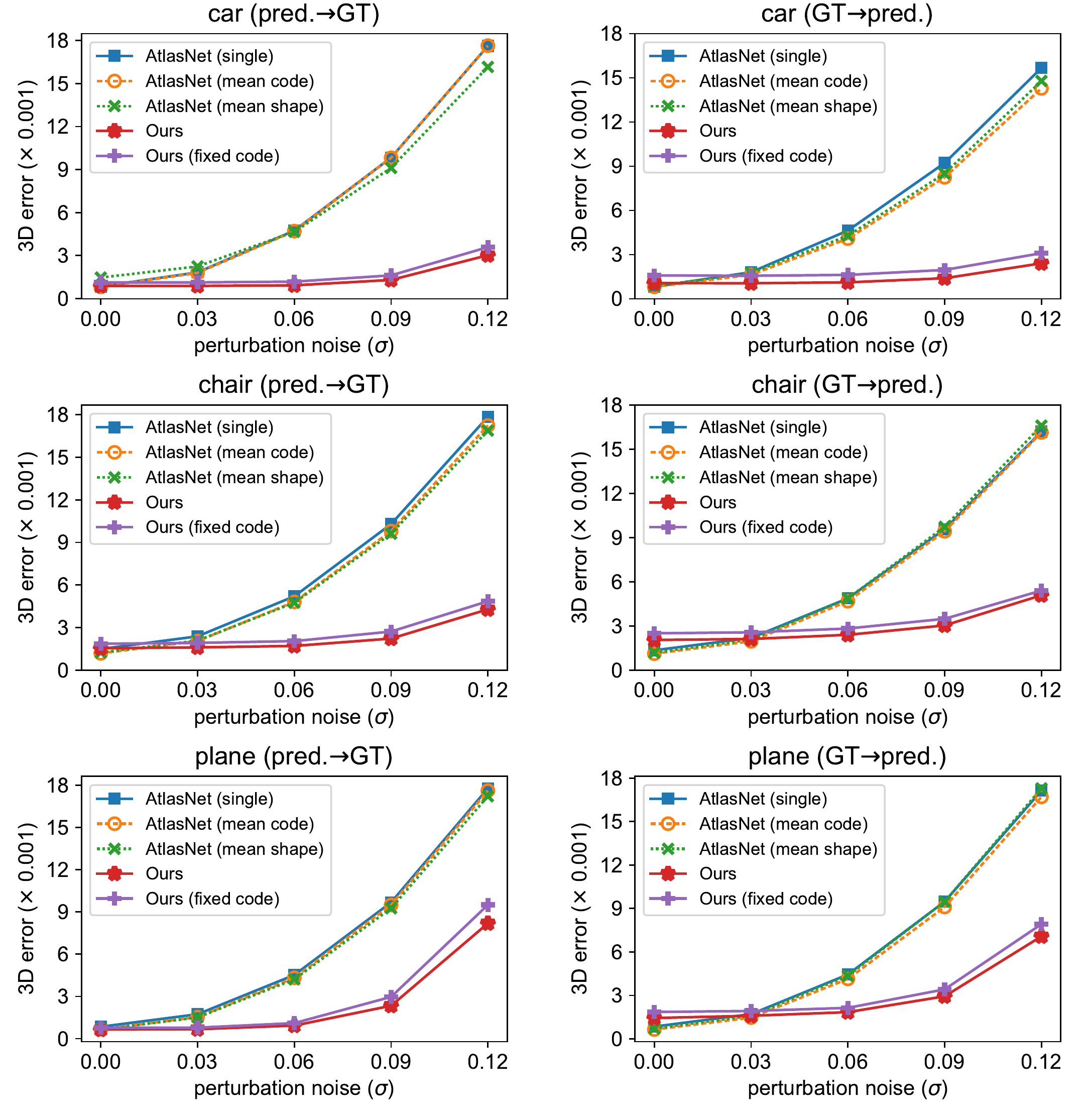}
	\caption{Category-specific performance to noise in coordinate system mapping.
	Our method is able to resolve for various extents of mesh misalignments from the sequence.
	}
	\label{fig:single_noise}
\end{figure}

We start by evaluating our mesh alignment in a category-specific setting.
We select the car, chair, and plane categories from ShapeNet, consisting of 703, 1356, and 809 objects in our test split, respectively.
For each object, we create an RGB sequence by overlaying its rendering onto a randomly paired SUN360 scene with the cameras in correspondence.
We retrain each category-specific AtlasNet model on our new dataset using the default settings for 500 epochs.
During optimization, we use the Adam optimizer~\cite{kingma2014adam} with a constant learning rate of $0.003$ for 100 iterations.
We manually set the penalty factors to be $\lambda_\text{code} = 0.05$ and $\lambda_\text{scale} = 0.02$.

One challenge is that the coordinate system for a mesh generated by AtlasNet is independent of the recovered world cameras $\{\cam_f\}$ for a real-world sequence.
Determining such coordinate system mapping (defined by a 3D similarity transform) is required to relate the predicted mesh to the world.
On the other hand, for the synthetic sequences, we know the exact mapping as we can render the views for AtlasNet and the input views $\{\I_f\}$ in the same coordinate system.

For our first experiment, we simulate the possibly incorrect mapping estimates by perturbing the ground-truth 3D similarity transform by adding Gaussian noise $\varepsilon \sim \mathcal{N}(0,\sigma\mathbf{I})$ to its parameters, pre-generated per sequence for evaluation.
We evaluate the 3D error metrics under such perturbations.
Note that our method utilizes no additional information other than the RGB information from the given sequences.

We compare our mesh reconstruction approach against three baseline variants of AtlasNet: (a) mesh generations from a single-image feed-forward initialization, (b) generation from the mean latent code averaged over all frames in the sequence, and (c) the mean shape where vertices are averaged from the mesh generation across all frames.

We show qualitative results in Fig.~\ref{fig:results_single} (compared under perturbation $\sigma=0.12$).
Our method is able to take advantage of multi-view geometry to resolve large misalignments and optimize for more accurate shapes.
The high photometric error from the background between views discourages mesh vertices from staying in such regions.
This error serves as a natural force to constrain the mesh within the desired 3D regions, eliminating the need of depth or mask constraints during optimization.
We further visualize our mesh reconstruction with textures that are estimated from all images (Fig.~\ref{fig:single_mesh}).
Note that the fidelity of mean textures increases while variance in textures decrease after optimization.

We evaluate quantitatively in Fig.~\ref{fig:single_noise}, where we plot the average 3D error over mapping noise.
This result demonstrates how our method handles inaccurate coordinate system mappings to successfully match the meshes against RGB sequences.
We also ablate optimizing the latent code $\z$, showing that allowing shape deformation improves reconstruction quality over a sole 3D similarity transform (``fixed code'' in Fig.~\ref{fig:single_noise}).
Note that our method is slightly worse in shape coverage error (GT$\to$pred.)\ when evaluated at the ground-truth mapping.
This result is attributed to the limitation of photometric optimization that  opts for degenerate solutions when objects are insufficiently textured.


\subsection{Multiple Object Categories} \label{sec:multi}

\begin{figure}[t!]
	\centering
	\includegraphics[width=1.0\linewidth]{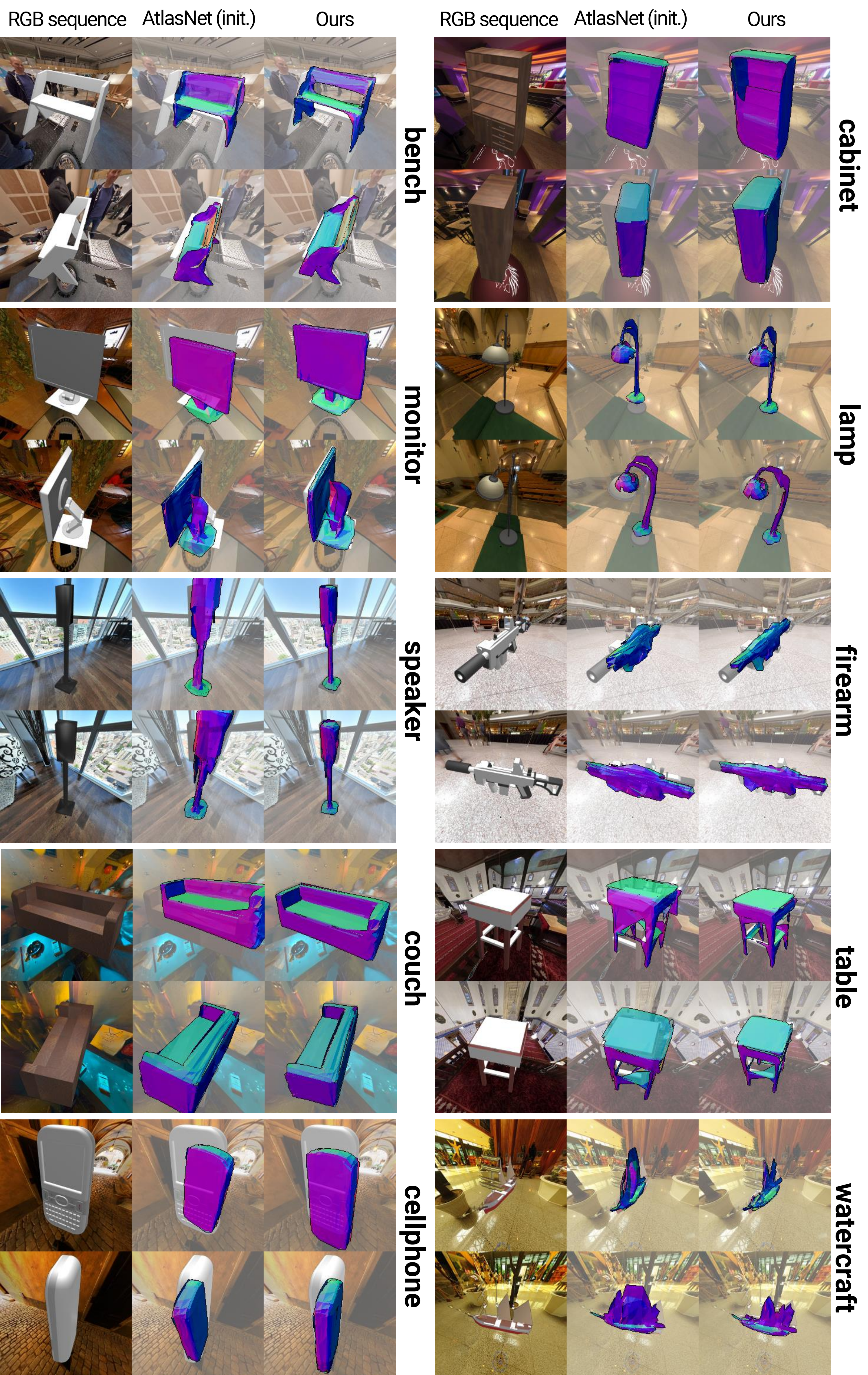}
	\caption{Qualitative results for general object categories.
	Our optimization method recovers subtle details such as back of benches, watercraft sails, and even starts to reveal cabinet open spaces which were initially occluded.
	Our method tends to fail more frequently with textureless objects (\eg, cellphone and firearm).
	\vspace{14pt}
	}
	\label{fig:results_multi}
\end{figure}

\begin{table*}[t!]
	\begin{subtable}{1\linewidth}
		\centering
		\setlength\tabcolsep{4pt}
		\resizebox{\linewidth}{!}{
		\begin{tabular}{l|c c c c c c c c c c c c c|c}
		    \toprule
			\bf Category & plane & bench & cabin. & car & chair & monit. & lamp & speak. & fire. & couch & table & cell. & water. & \bf mean \\
			\midrule
			AtlasNet (single) & {3.872}&{4.931}&{5.708}&{4.269}&{4.869}&{4.687}&{8.684}&{7.245}&{3.864}&{5.017}&{4.964}&{4.571}&{4.290}&{5.152}\\ 
			AtlasNet (mean code) & {3.746}&{4.496}&{5.600}&{4.286}&{4.571}&{4.634}&{7.366}&{6.976}&{3.632}&{4.798}&{4.903}&{4.286}&{3.860}&{4.858}\\ 
			AtlasNet (mean shape) & {3.659}&{4.412}&{5.382}&{4.192}&{4.499}&{4.424}&{\bf 7.200}&{6.683}&{3.547}&{4.606}&{4.860}&{4.196}&{3.742}&{4.723}\\ 
			Ours & \bf 0.704&\bf 1.821&\bf 2.850&\bf 0.597&\bf 1.441&\bf 1.115&8.855&\bf 3.430&\bf 1.255&\bf 0.983&\bf 1.725&\bf 1.599&\bf 1.743&\bf 2.163\\
			\bottomrule 
		\end{tabular}
		}
		\caption{3D error: prediction $\to$ ground truth (shape accuracy).}
	\end{subtable}
	\begin{subtable}{1\linewidth}
		\centering
		\setlength\tabcolsep{4pt}
		\resizebox{\linewidth}{!}{
		\begin{tabular}{l|c c c c c c c c c c c c c|c}
		    \toprule
			\bf Category & plane & bench & cabin. & car & chair & monit. & lamp & speak. & fire. & couch & table & cell. & water. & \bf mean \\
			\midrule
		
			AtlasNet (single) & {4.430}&{4.895}&{5.024}&{4.461}&{4.896}&{4.640}&{8.906}&{6.994}&{4.407}&{4.613}&{5.350}&{4.254}&{4.263}&{5.164}\\ 
			AtlasNet (mean code) & {4.177}&{4.507}&{4.962}&{4.384}&{4.635}&{4.143}&{\bf 7.292}&{6.990}&{4.307}&{4.463}&{\bf 5.084}&{4.036}&{3.718}&{4.823}\\ 
			AtlasNet (mean shape) & {4.464}&{4.915}&{5.150}&{4.521}&{4.940}&{4.560}&{8.159}&{7.308}&{4.528}&{4.707}&{5.255}&{4.299}&{4.183}&{5.153}\\ 
			Ours & \bf 2.237&\bf 3.215&\bf 1.927&\bf 0.734&\bf 2.377&\bf 2.119&10.764&\bf 4.152&\bf 2.583&\bf 1.735&6.126&\bf 1.851&\bf 2.926&\bf 3.288\\ 
			\bottomrule
		\end{tabular}
		}
		\caption{3D error: ground truth $\to$ prediction (shape coverage).}
	\end{subtable}
	\caption{Average 3D test error for general object categories (numbers scaled by $10^3$).
	The mean is taken across categories.
	Our optimization method is effective on most object categories.
	Note that our method improves on accuracy of the table category despite worsening in shape coverage due to insufficient textures in object samples.}
	\label{table:multicat}
\end{table*}

We extend beyond a model that reconstructs a single object category by training a single model to reconstruct multiple object categories.
We take 13 commonly chosen CAD model categories from ShapeNet~\cite{choy20163d,fan2017point,groueix2018atlasnet,lin2018learning}. 
We follow the same settings as in Sec.~\ref{sec:single} except we retrain AtlasNet longer for 1000 epochs due to a larger training set.

We show visual results in Fig.~\ref{fig:results_multi} on the efficacy of our method for multiple object categories (under perturbation $\sigma=0.12$).
Our results show how we can reconstruct a shape that better matches our RGB observations (e.g., refining hollow regions, as in the bench backs and table legs).
We also show category-wise quantitative results in Table~\ref{table:multicat}, compared under perturbation noise $\sigma=0.06$.
We find photometric optimization to perform effectively across most categories except lamps, which consist of many examples where optimizing for thin structures is hard  for photometric loss.


\subsection{Real-world Videos} \label{sec:real}

Finally, we demonstrate the efficacy of our method on challenging real-world video sequences orbiting an object.
We use a dataset of RGB-D object scans~\cite{choi2016large}, where we use the chair model to evaluate on the chair category.
We select the subset of video sequences that are 3D-reconstructible using traditional pipelines~\cite{schoenberger2016sfm} and where \SFM extracts at least 20 reliable frames and 100 salient 3D points.
We retain 82 sequences with sufficient quality for evaluation.
We rescale the sequences to $240\times 320$ and skip every 10 frames.

\begin{figure}[t!]
	\centering
	\includegraphics[width=1.0\linewidth]{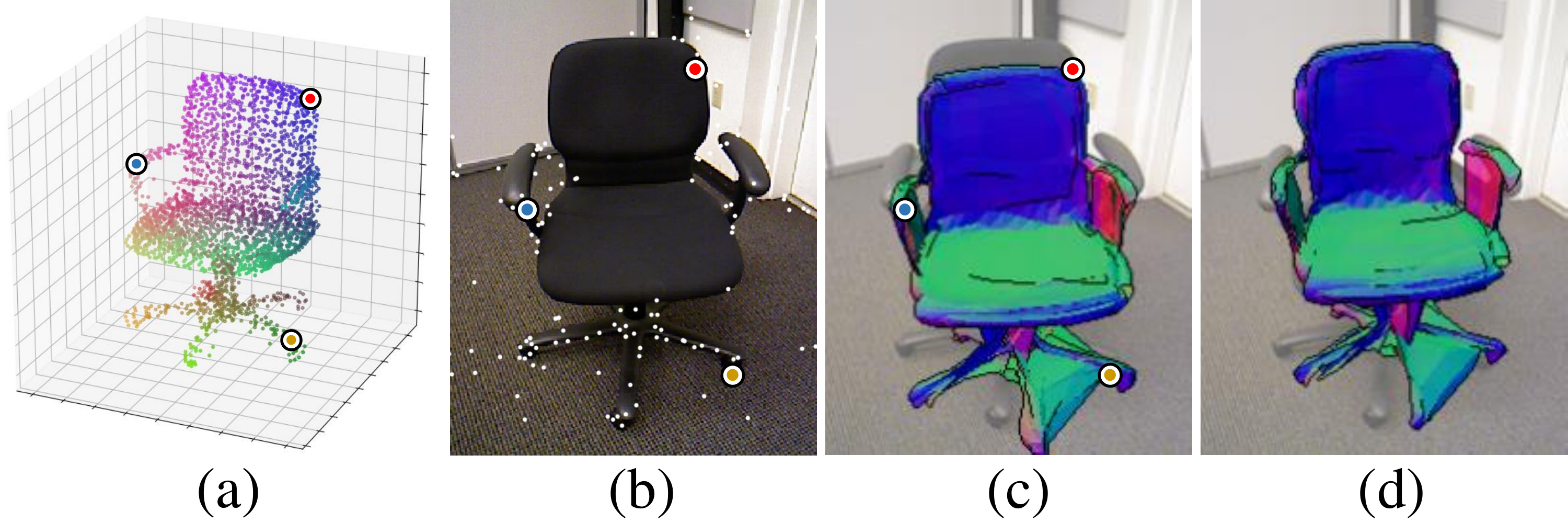}
	\caption{We select 3 correspondences between (a) the mesh vertices and (b) the \SFM points to find (c) an estimated coordinate system mapping by fitting a 3D similarity transform.
	(d) Alignment result after our photometric optimization.}
	\label{fig:annotate}
\end{figure}

We compute the camera extrinsic and intrinsic matrices using off-the-shelf \SFM with COLMAP~\cite{schoenberger2016sfm}.
For evaluation, we additionally compute a rough estimate of the coordinate system mapping by annotating 3 corresponding points between the predicted mesh and the sparse points extracted from \SFM (Fig.~\ref{fig:annotate}), which allows us to fit a 3D similarity transform.
We optimize using Adam with a learning rate of 2e-3 for 200 iterations, and we manually set the penalty factors to be $\lambda_\text{code} = 0.05$ and $\lambda_\text{scale} = 0.01$.

\begin{figure*}[t!]
	\centering
	\includegraphics[width=1.0\linewidth]{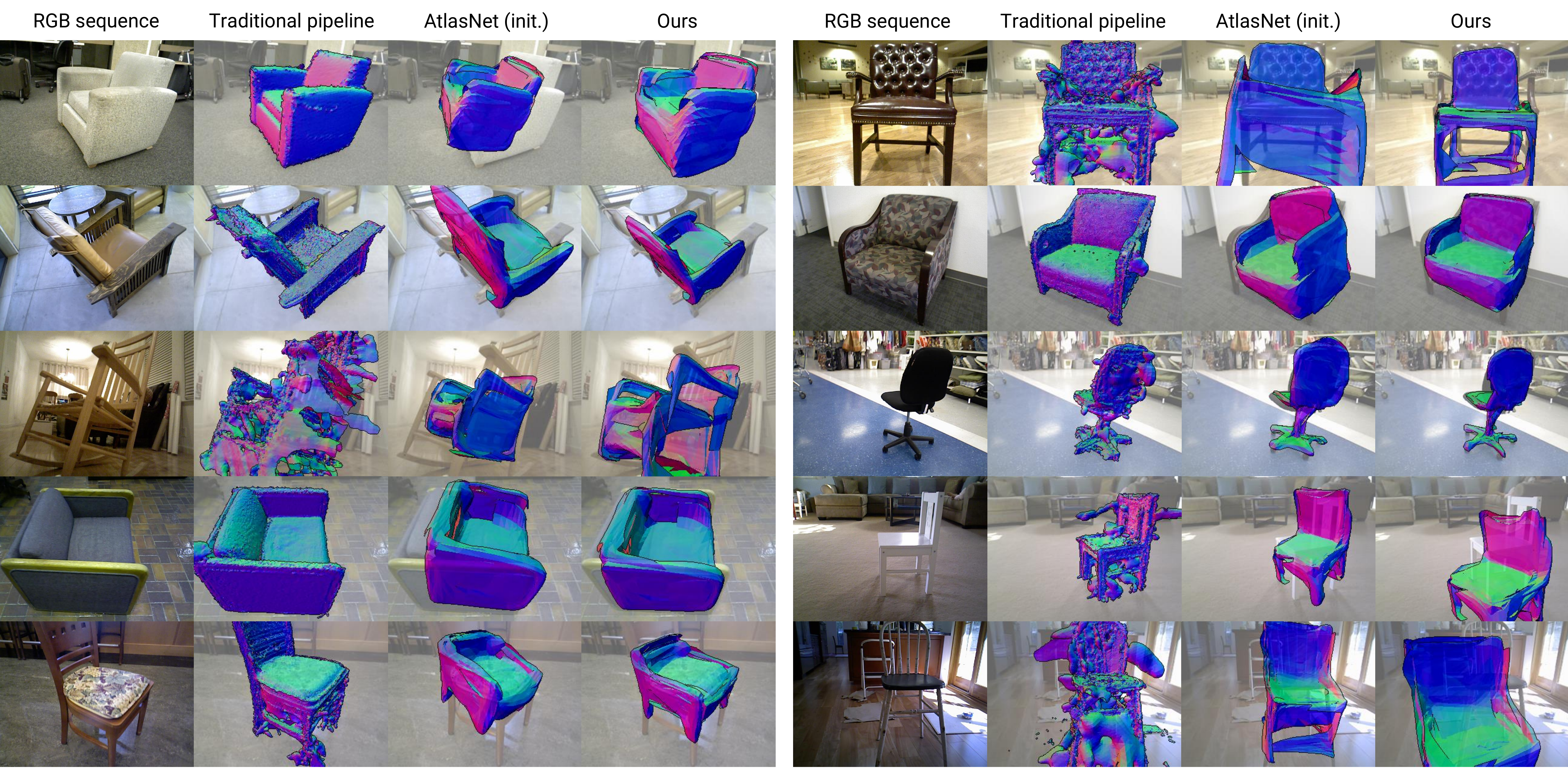}
	\caption{Qualitative results on real-world sequences.
	Given an initialization, our method accurately aligns a generated mesh to an RGB video.
	Even when the initial mesh is an inaccurate prediction of the real object, our method is still able to align the semantic parts (bottom left).
	We show failure cases in the last two examples in the bottom right, where there is insufficient background texture as photometric cues and where the initial mesh is insufficient to capture the thin structures.
	We also show the result of a traditional reconstruction pipeline~\cite{schoenberger2016sfm} after manual cleanup. Due to the difficulty of the problem these meshes still often have many  undesirable artifacts.
	}
	\label{fig:results_koltun}
\end{figure*}

\begin{figure}[t!]
	\begin{floatrow}
		\centering
		\capbtabbox{
			\begin{tabular}{c|c c}
				Dist. & Initial. & Optim. \\
				\midrule
				1 & 6.504 & 4.990 \\ 
				2 & 9.064 & 6.979 \\
				3 & 10.984 & 8.528 \\
				4 & 12.479 & 9.788 \\
				6 & 14.718 & 11.665 \\
			\end{tabular}
		}{\caption{Average pixel reprojection error (scaled by $100$) from real-world videos as a function of frame distances.}\label{table:koltun_reproj}}
		\hspace{-15pt}
		\ffigbox{
			\includegraphics[width=1.0\linewidth]{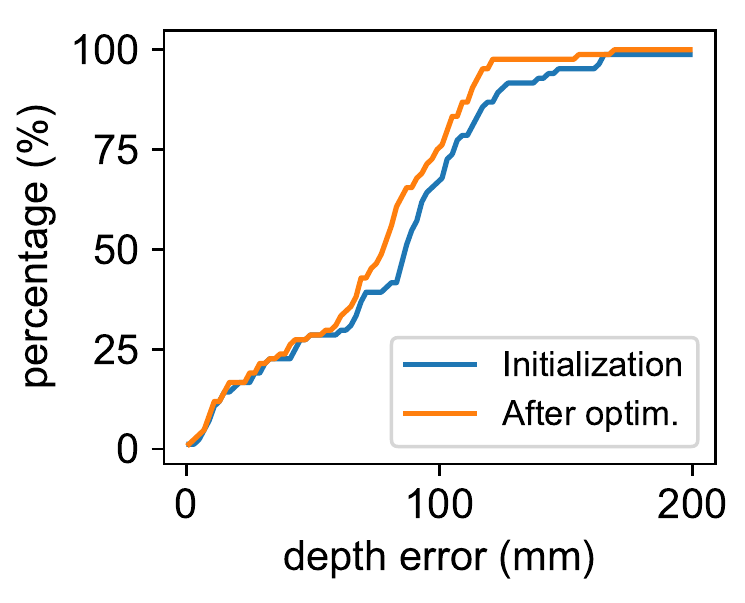}
		}{\caption{Metric-scale depth error before and after optimization (with \SFM world cameras rescaled).}\label{fig:koltun_depth}}
	\end{floatrow}
\end{figure}

We demonstrate how our method is applicable to real-world datasets in Fig.~\ref{fig:results_koltun}.
Our method is able to refine shapes such as armrests and office chair legs.
Note that our method is sensitive to the quality of mesh initialization from real images, mainly due to the domain mismatch between synthetic and real data during the training/test phases of the shape prior.
Despite this, it is still able to straighten and align to the desired 3D location.
In addition, we report the average pixel reprojection error in Table~\ref{table:koltun_reproj} and metric depth error in Fig.~\ref{fig:koltun_depth} to quantify the effect of photometric optimization, which shows further improvement over coarse initializations.

Finally, we note that surface reconstruction is a challenging post-processing procedure for traditional pipelines. 
Fig.~\ref{fig:results_koltun} shows sample results for  \SFM~\cite{schoenberger2016sfm}, PatchMatch Stereo~\cite{bleyer2011patchmatch}, stereo fusion, and Poisson mesh reconstruction~\cite{kazhdan2013screened} from COLMAP~\cite{schoenberger2016sfm}.
In addition to the need of accurate object segmentation, the dense meshing problem with traditional pipelines typically yields noisy results without laborious manual post-processing.

\section{Conclusion}

We have demonstrated a method for reconstructing a 3D mesh from an RGB video by combining data-driven deep shape priors with multi-view photometric consistency optimization. 
We also show that mesh rasterization from a virtual viewpoint is critical for avoiding degenerate photometric gradients during optimization.
We believe our photometric mesh optimization technique has merit for a number of practical applications.
It enables the ability to generate more accurate models of real-world objects for computer graphics and potentially allows automated object segmentation from video data.
It could also benefit 3D localization for robot navigation and autonomous driving, where accurate object location, orientation, and shape from real-world cameras is crucial for more efficient understanding.

{\small
\bibliographystyle{ieee_fullname}
\bibliography{reference}
}

\newpage
\clearpage
\section*{A. Appendix}

\subsection*{A.1. Architectural and Pretraining Details}

We use AtlasNet~\cite{groueix2018atlasnet} as the base network architecture for our experiments.
Following Groueix~\etal~\cite{groueix2018atlasnet}, the image encoder is the ResNet-18~\cite{he2016deep} architecture where the last fully-connected layer is replaced with one with an output dimension of 1024, which is the size of the latent code.
We use the 25-patch version of the AtlasNet mesh decoder, where each deformable patch is an open triangular mesh with $5^2 \times 2 = 50$ triangles on a $5\times 5$ regular grid. We redirect the readers to Groueix~\etal~\cite{groueix2018atlasnet} for more details.

In the stage of pretraining AtlasNet on ShapeNet~\cite{chang2015shapenet} with textured background from SUN360~\cite{xiao2012recognizing}, we train all networks using the Adam optimizer~\cite{kingma2014adam} with a constant learning rate of $10^{-4}$.
We set the batch size for all experiments to be 32.
We initialize the AtlasNet encoder with the pretrained ResNet-18 on ImageNet~\cite{russakovsky2015imagenet} except for the last modified layer (before the latent code), and we initialize the decoder with that pretrained from a point cloud autoencoder from Groueix~\etal~\cite{groueix2018atlasnet}.

\subsection*{A.2. Warp Parameterization Details}
We parameterize the rotation component of 3D similarity transformations with the $\mathfrak{so}(3)$ Lie algebra.
Given a warp parameter vector $\bomega = [\omega_1,\omega_2,\omega_3]^\top \in \mathfrak{so}(3)$, the rotation matrix $\Rot(\bomega) \in \mathbb{SO}(3)$ can be written as
\begin{equation}
\Rot(\bomega) = \exp \left( 
\begin{bmatrix} 0 & -\omega_3 & \omega_2 \\ \omega_3 & 0 & -\omega_1 \\ -\omega_2 & \omega_1 & 0 \end{bmatrix}
\right) \;,
\end{equation}
where $\exp$ is the exponential map (\ie matrix exponential).
$\Rot$ is the identity transformation when $\bomega$ is an all-zeros vector.
The exponential map is Taylor-expandable as
\begin{equation}
\Rot(\bomega) = \exp(\bomega_\times) = \lim_{K \to \infty} \sum_{k=0}^K \frac{\bomega_\times^k}{k!} \;.
\end{equation}
We implement the $\mathfrak{so}(3)$ parameterization using the Taylor approximation expression with $K=20$.
We have also tried parametrizing the 3D similarity transformations with the self-contained Lie group $\text{Sim}(3)$, where the scale is incorporated into the exponential map; we find it to yield almost identical results.
We also take the exponential on the scale $s$ to ensure positivityl; the resulting scale does not change when $s=0$.

\subsection*{A.3. SUN360 Background Data Generation}

\begin{figure}[t!]
	\centering
	\includegraphics[width=1.0\linewidth]{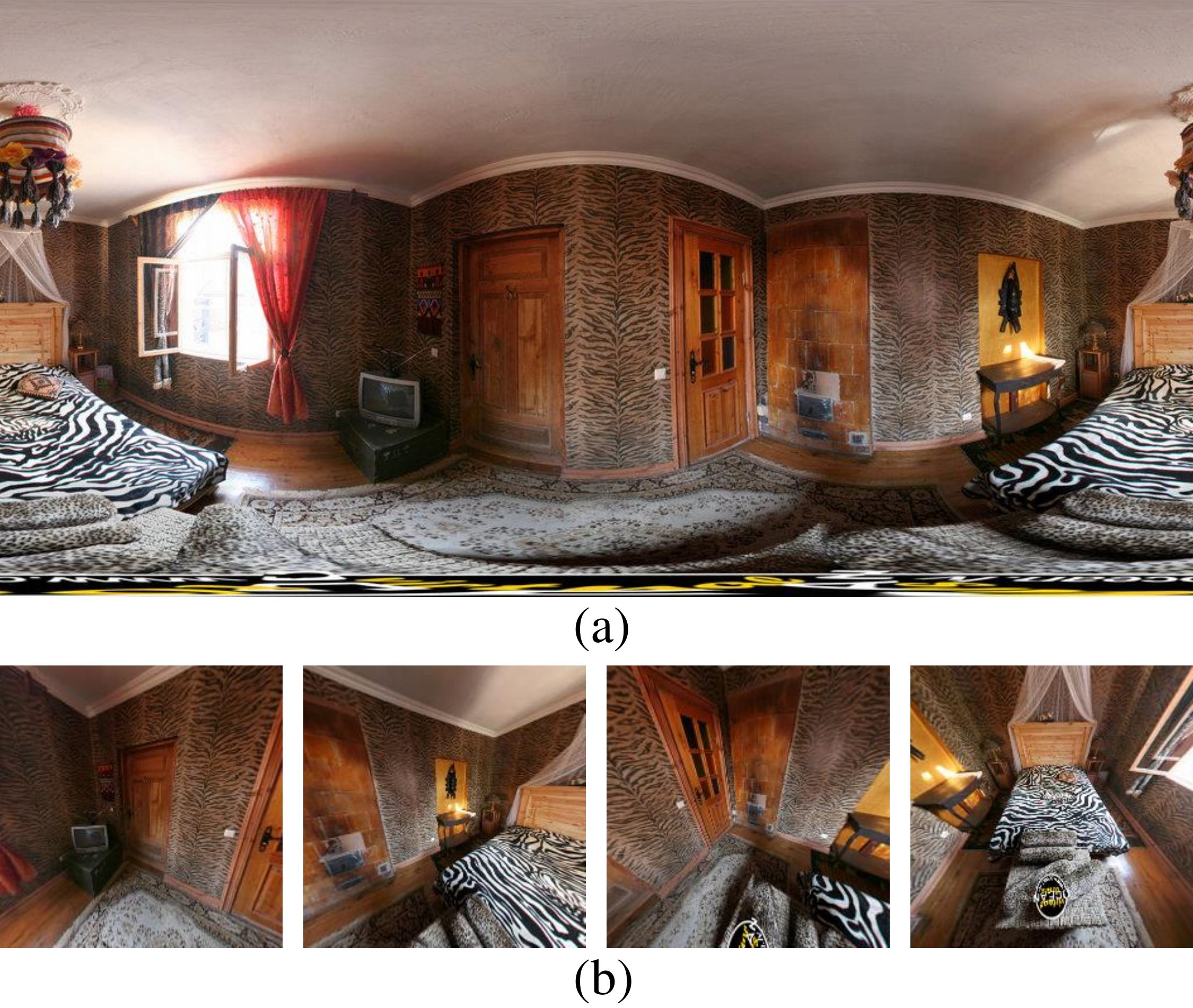}
	\caption{(a) Example panoramic (spherical) image and (b) sample cropped images at different camera viewpoints.
	}
	\label{fig:sun360}
\end{figure}

The background images from SUN360~\cite{xiao2012recognizing} are cropped from spherical images with a resolution of 1024$\times$512, using a field of view of 90$\degree$.
Fig.~\ref{fig:sun360} illustrates an example of the original spherical image and its generated crops.

\end{document}